\documentclass[sigconf]{acmart}

\AtBeginDocument{%
  }

\copyrightyear{2026}
\acmYear{2026}
\setcopyright{cc}
\setcctype{by}
\acmConference[ICMR '26]{International Conference on Multimedia Retrieval}{June 16--19, 2026}{Amsterdam, Netherlands}
\acmBooktitle{International Conference on Multimedia Retrieval (ICMR '26), June 16--19, 2026, Amsterdam, Netherlands}
\acmDOI{10.1145/3805622.3810692}
\acmISBN{979-8-4007-2617-0/2026/06}

\begin{document}

\title{Object Referring-Guided Scanpath Prediction with Perception-Enhanced Vision-Language Models}


\author{Rong Quan}
\authornote{Both authors contributed equally to this research.}
\authornote{Corresponding author. This work is supported in part by the National Natural Science Foundation of China under grant 62272229, 62206127, and U2441285.}
\affiliation{%
 \institution{Nanjing University of Aeronautics and Astronautics}
  \city{Nanjing}
  \country{China}}
\email{rongquan0806@gmail.com}

 \author{Yantao Lai}
 \authornotemark[1]
 \affiliation{%
 \institution{Nanjing University of Aeronautics and Astronautics}
  \city{Nanjing}
  \country{China}}
 \email{yantaolai@nuaa.edu.cn}

\author{Dong Liang}
\affiliation{%
 \institution{Nanjing University of Aeronautics and Astronautics}
  \city{Nanjing}
  \country{China}}
\email{liangdong@nuaa.edu.cn}

 \author{Jie Qin}
 \affiliation{%
 \institution{Nanjing University of Aeronautics and Astronautics}
  \city{Nanjing}
  \country{China}}
 \email{qinjiebuaa@gmail.com}

\renewcommand{\shortauthors}{Rong Quan, Yantao Lai, Dong Liang, Jie Qin}

\begin{abstract}
  Object Referring-guided Scanpath Prediction (ORSP) aims to predict the human attention scanpath when they search for a specific target object in a visual scene according to a linguistic description describing the object. Multimodal information fusion is a key point of ORSP. Therefore, we propose a novel model, ScanVLA, to first exploit a Vision-Language Model (VLM) to extract and fuse inherently aligned visual and linguistic feature representations from the input image and referring expression.
Next, to enhance the ScanVLA’s perception of fine‑grained positional information, we not only propose a novel History Enhanced Scanpath Decoder (HESD) that directly takes historical fixations’ position information as input to help predict a more reasonable position for the current fixation, but also adopt a frozen Segmentation LoRA as an auxiliary component to help localize the referred object more precisely, which improves the scanpath prediction task without incurring additional large computational and time costs.
Extensive experimental results demonstrate that ScanVLA can significantly outperform existing scanpath prediction methods under object referring. 
\end{abstract}




\maketitle
\section{Introduction}
Object referring~\cite{vasudevan2018object} aims to localize a specific target object in a visual scene based on a given linguistic description describing the object. 
Object referring devotes to imitating human beings' extraordinary capability of simultaneously handling multimodal information, which has important application in human-computer interaction (HCI)~\cite{bansal2024hoi}, robotics\cite{gao2020visual}, image retrieval\cite{hoiem2004object}, and so on. 
Gaze plays a crucial role in object referring. 
Over the past few decades, extensive research has demonstrated the very tight link between a word in a referring expression and the next eye movements of the person hearing it~\cite{pixel,lookhear,connecting,TSPT}, suggesting that humans always incrementally integrate visual information and word-by-word linguistic guidance during the object referring process. Human scanpath~\cite{pathformer3d,VSPT}, formed by sequential human eye fixations when exploring a visual scene, can reflect the dynamic human attention shifting process.
Considering modern human beings' pursuit of increasingly intelligent HCI functions such as fully automatic unmanned driving and humanoid robots, we investigate human scanpath prediction during the object referring process 
in this paper, to realize a more in-depth and comprehensive imitation of human beings' dynamic attention mechanism when searching a referred object in a visual scene.  
Looking ahead, ORSP will have extensive applications in various HCI fields, such as robotics~\cite{aronson2022gaze, saran2020understanding}, autonomous driving~\cite{driver} and VR/AR~\cite{lavoie2024comparing, pai2016gazesim}.

\begin{figure}[t]
\centering
\includegraphics[width=1\columnwidth]{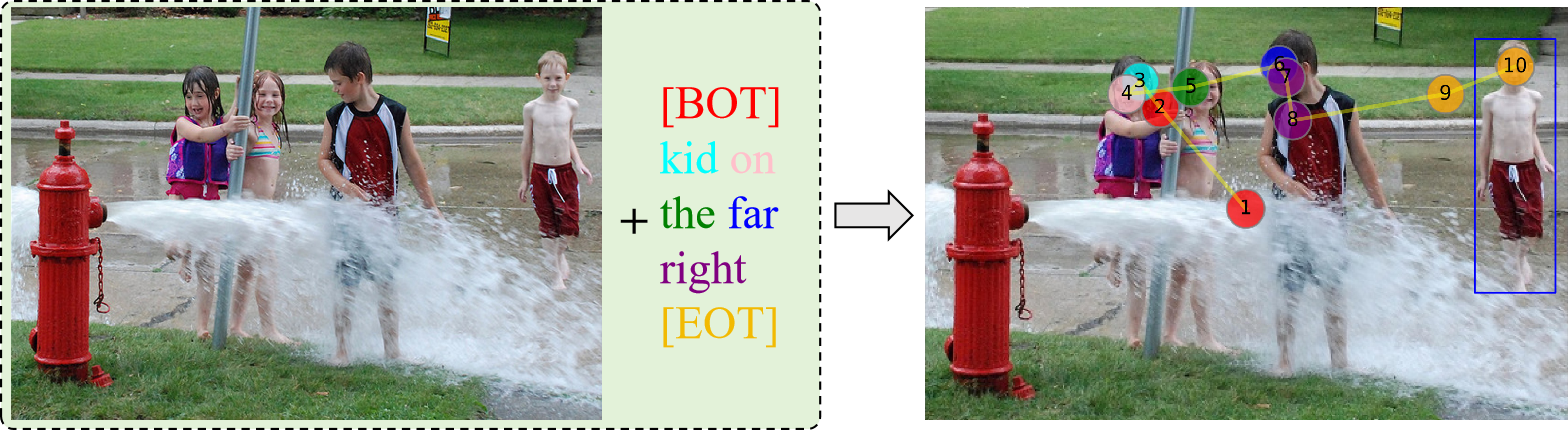} 
\caption{\textbf{Illustration of Object Referring-guided Scanpath Prediction task}. The blue bounding boxes denote the referred object. 
Each word in the referring expression corresponds to a fixation pack (represented by numbered circles, where numbers denote fixation order) of the same color.
As the model incrementally receives each word, it is required to predict the corresponding fixation pack.}
\label{fig: intro}
\vspace{-8pt}
\end{figure}

As shown in Figure \ref{fig: intro}, different from traditional scanpath predction tasks\cite{cartella2025modeling,kara2025diffeye,lee2025streamgaze,scandmm,scangan360}, ORSP involves incrementally predicting human eye movements as they search for a specific target in an image while listening to the corresponding referring expression that describes the target. 
The only substantial work on this task is ART~\cite{lookhear}, which employs an autoregressive Transformer decoder~\cite{transformer} to predict, for each word, a variable number of fixations based on the fixation history. 
Although achieving quite good performance, ART still has some inevitable issues. For example, a key point of ORSP is multimodal information fusion, \emph{i.e.,} the matching and fusion of the visual and linguistic information. ART employs separate visual and semantic encoders (ResNet-50~\cite{resnet50} and RoBERTa~\cite{roberta}) for feature extraction, resulting in significant differences between the extracted features, which poses greater difficulties for subsequent fusion. Although they attempted to fine-tune the parameters of the visual and semantic encoders, due to the size limitation of training data, the fine-tuned models still had limitations in extracting mutually matching visual and text features.
Moreover, ART adopts two auxiliary tasks, including object localization and target category prediction, to enhance its scanpath prediction performance. Although pre-training and training the auxiliary tasks can enhance performance to some extent, huge computational burden and long computation time are coming together.

\begin{figure*}[t]
\centering
\includegraphics[width=0.8\textwidth]{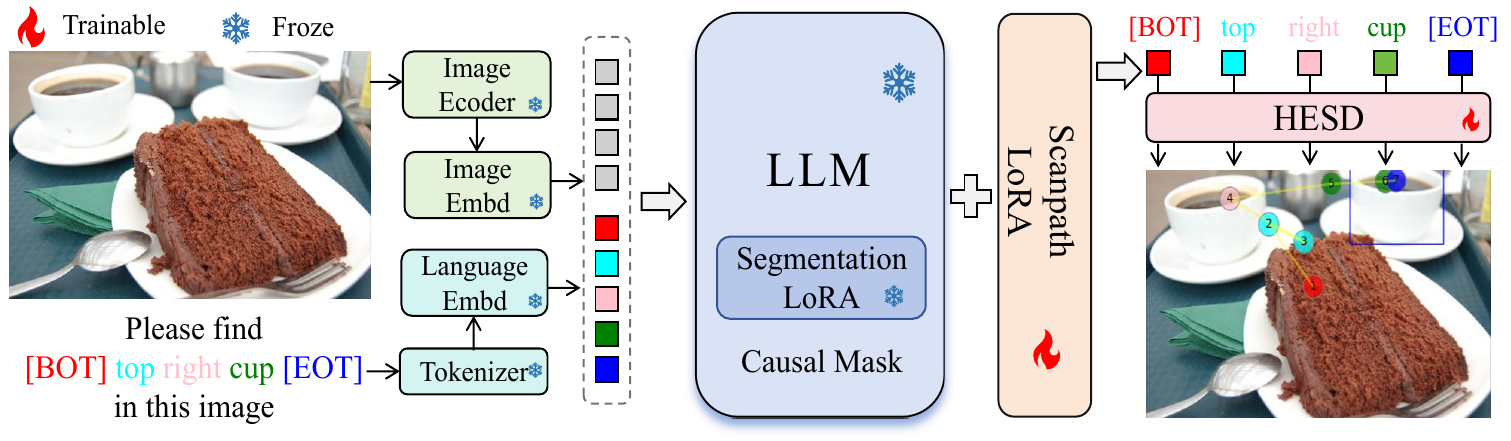} 
\vspace{-6pt}
\caption{\textbf{Overall architecture of ScanVLA }. For each word in the referential expression, 
ScanVLA uses a Tokenizer, Image Encoder, and embedders (for image/text) to get joint embeddings. These embeddings are fed into an autoregressive LLM with causal mask for fusion, 
then forwarded to History Enhanced Scanpath Decoder (HESD) to produce fixation sequences.}
\label{fig: ScanVLA}
\vspace{-6pt}
\end{figure*}

Recent years have witnessed the rapid development of Vision-Language Models (VLMs)~\cite{internvl25, Qwen3-VL, qwenvl25}. Through training on extremely large vision-language datasets, VLM can obtain human-level performance in mutimodal perception. Therefore, in this paper, 
we propose a novel framework named ScanVLA, which exploits a VLM for multimodal feature extraction and fusion in ORSP. By resorting to the powerful understanding and integration capabilities of VLM for visual and linguistic information, ScanVLA can first extract inherently aligned visual-linguistic representations from the input image and linguistic description. In addition, we adopt LoRA (Low-Rank Adaptation)~\cite{lora} to selectively fine-tune task-specific parameters of VLM\cite{internvl25, Qwen3-VL} while preserving the model's original perception capabilities. 

After obtaining inherently aligned visual-linguistic features, we further propose a novel History Enhanced Scanpath Decoder (HESD). Instead of exploiting a latent state that integrates all the historical fixations' information, HESD directly inputs the historical fixations' fine-grained positions to help predict current fixation's position, which results in more reasonable fixation positions. 
Moreover, unlike ART which first pre-trains the auxiliary tasks, and then simultaneously trains the auxiliary tasks during the model training process, to help improve its final scanpath prediction performance. Here in our work, we integrate a frozen Segmentation LoRA — analogous to using a plug-and-play plugin — as the auxiliary component, to help improve the scanpath prediction performance without devoting substantial calculation and time consumption.

In summary, our work makes three major contributions:
\vspace{-3pt}
\begin{itemize}
 \item We propose ScanVLA, a novel Vision-Language-Action (VLA) Model tailored for the ORSP task. ScanVLA is the first to employ a VLM as both encoding and fusion module for image and text, effectively addressing the feature alignment issues encountered by traditional scanpath prediction methods.
 \item To enhance fine-grained perception capabilities required for the ORSP task, ScanVLA not only proposes a novel HESD to effectively fuse historical fixation information, but also employs a frozen Segmentation LoRA, pre-trained on various segmentation-related tasks, as an auxiliary module.
 \item Extensive experimental results demonstrate that our model significantly outperforms existing methods across multiple metrics, validating the inherent advantages of ScanVLA.
\end{itemize}

\section{Methods}
\subsection{Preliminaries}
Given an input image $I \in \mathbb{R}^{3 \times H \times W}$ and a referential expression $R_{ref} = \{w_1, \ldots, w_L\}$ containing $L$ words, the ORSP task  predicts a pixel-level fixation pack $\mathcal{P}_j$ for each word $w_j$, where $\mathcal{P}_j = \{ (x_i^j, y_i^j) \mid i = 0, 1, \ldots \}$ denotes an ordered sequence of 2D fixations related to $w_j$. 
Unlike prior works~\cite{glamm}, which primarily focus on grounding noun phrases, our approach explicitly incorporates relational terms (e.g., ‘on,’ ‘under’) into the learning process, ensuring that the model can better capture spatial and semantic relationships.

Additionally, another distinctive characteristic of our task is its incremental nature, which means that when predicting the fixation pack $\mathcal{P}_j$ corresponding to the word $w_j$, only the prefix text $\{w_1,...,w_j\}$ and image information $I \in \mathbb{R}^{3 \times H \times W}$ can be utilized. After obtaining the fixation pack for each word, we concatenate all predicted fixation packs together to form a complete scanpath $\mathcal{S} = \{\mathcal{P}_0, \mathcal{P}_1,...,\mathcal{P}_{L+1}\}$. 
It is worth noting that we follow the setting of ART\cite{lookhear} to append the special tokens \texttt{[BOT]} (Beginning-Of-Text) and \texttt{[EOT]} (End-Of-Text) before and after $R_{ref}$, respectively, to help predict the initial fixation pack $\mathcal{P}_0$ (triggered before the first word is spoken) and the final fixation pack $\mathcal{P}_{L+1}$ (triggered after the target referential expression ends). This transforms $R_{ref}$ into $R = \{\texttt{[BOT]}, w_1, \ldots, w_L, \texttt{[EOT]}\}$.

\subsection{Architecture}
\subsubsection{Feature Extraction and Fusion}
We employ pre-trained LLaVA-like\cite{llava} models (here, InternVL 2.5~\cite{internvl25} 1B and Qwen3-VL~\cite{Qwen3-VL} 2B) as the feature extraction and fusion module, which consists of a visual encoder, a tokenizer and visual/language Embd layers.
Using these components, when predicting the fixation pack for the \(w_j\)-th word,  we obtain the hidden states $\mathcal{H}_j$ corresponding to the pack $\mathcal{P}_j$.
For a referential expression with a length of $L+2$ (including \texttt{[BOT]} and \texttt{[EOT]}), the hidden state of $\mathcal{H}_j, j\in [0,1,..,L+1]$ can be obtained.

We leverage pre-trained VLMs while addressing their limitations through two key innovations: 1) Incorporating a frozen `Segmentation LoRA' (from Sa2VA~\cite{sa2va}) to enhance the understanding of referential targets and fine-grained perception capabilities, compensating for VLMs’ weak visual grounding; 2) Maintaining a trainable `Scanpath LoRA' to learn knowledge relevant to the ORSP task. Furthermore, the use of these two LoRAs offers advantages such as plug-and-play functionality and lower computational costs. In addition, unlike conventional methods with bidirectional masks (e.g., BERT~\cite{BERT}, Ernie\cite{sun2019ernie}), our LLM employs a causal mask with an inverted triangular structure. This design is deliberately aligned with ORSP’s incremental processing characteristic, ensuring that only prefix text information could be seen.

\subsubsection{History Enhanced Scanpath Decoder}
\begin{figure}[t]
\centering
\includegraphics[width=0.8\columnwidth]{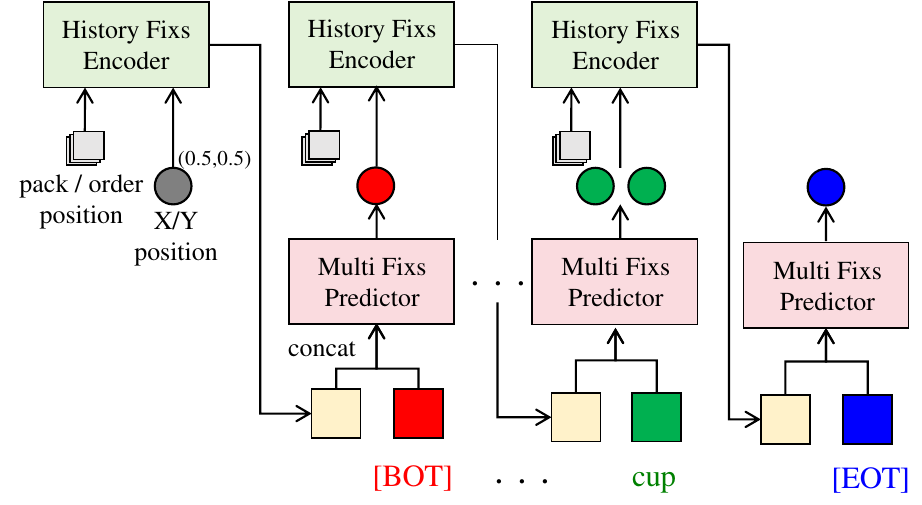} 
\caption{Detailed architecture of HESD.}
\label{fig: scanpath decoder}
\vspace{-10pt}
\end{figure}

The detailed architecture of the History Enhanced Scanpath Decoder (HESD) is illustrated in Figure \ref{fig: scanpath decoder}.
HESD first concatenate historical fixation coordinates (X/Y values), pack indices $pack$, and intra-pack fixation indices $order$ into a tensor $T \in R^{L_{h}\times4}$, where $L_{h}$ is the total number of historical fixations. 
This tensor is then encoded in the \textbf{History Fixs Encoder} to represent the global historical fixation information. For simplicity and efficiency, we implement this component using a unidirectional GRU~\cite{gru}; alternative architectures such as the Transformer Encoder~\cite{transformer} or LSTM\cite{lstm} can also be adopted. The obtained global historical information and text embedding $\mathcal{H}_j$ can be concatenated to yield the final hidden state representation $\mathcal{H}_j'$
Finally, we employ the \textbf{Multi-Fixs Predictor} module to generate the corresponding fixation coordinates $X$, $Y$ and validity probability $V$ from each hidden state $\mathcal{H}_j', j\in [0,1,..,L+1]$. This predictor consists of three independent two-layer MLP structures.

To handle variable fixations per pack, we preprocess the data by first setting a maximum number of fixations per pack, $L_{\mathcal{P}}$, and truncating any pack exceeding this limit, then representing valid fixations with a fixed token $FIX$, and finally padding packs with fewer than $L_{\mathcal{P}}$ using a padding token $PAD$.
During generation, we conduct post-processing by traversing all fixations in each pack
$\mathcal{P} = \{ (x_i, y_i) \mid i = 0, 1, \ldots \}$ from $i=0$ to $i=L_p-1$. If the validity probability $V^i_j$ of the current fixation is bigger than or equal to 0.5, we add its position information $X^i_j$ and $Y^i_j$ to our predict scanpath. Otherwise, we stop the traversal of current pack. Finally, the predicted $\hat{\mathcal{P}_j}, j\in [0,1,..,L+1]$ can be obtained.


\subsection{Loss Function}
The model is trained end-to-end using the auxiliary text generation loss ${\mathcal{L}}_{txt}$, position loss ${\mathcal{L}}_{txt}$ and validity loss $\mathcal{L}_{token}$. The overall objective $\mathcal{L}_{total}$ is the sum of these losses:

\begin{equation} 
\begin{aligned}
\mathcal{L}_{total}&=\mathcal{L}_{txt}+\mathcal{L}_{xy}+\mathcal{L}_{token} \\
\mathcal{L}_{txt} &= \frac{1}{L_{txt}}\sum_{i=1}^{L_{txt}}CE(\hat{\mathbf{y}}_{i}, \mathbf{y}_{i}) \\
\mathcal{L}_{xy}&=\frac{1}{L_{txt}l}\sum_{i=1}^{L_{txt}}\sum_{j=1}^{l}\left(|x_{i,j}-\hat{x}_{i,j}|+|y_{i,j}-\hat{y}_{i,j}|\right) \\
\mathcal{L}_{token}&=\frac{1}{L_{txt}L_P}\sum_{i=1}^{L_{txt}}\sum_{j=1}^{L_{P}}(Focal Loss(v_{i,j}, \hat{v}_{i,j})) \\
\end{aligned}
\end{equation}

Specifically, ${\mathcal{L}}_{txt}$ is the auxiliary auto-regressive cross-entropy ($CE$) loss and $\mathcal{L}_{xy}$ adopts the $L1$ regression loss.
The purpose of $\mathcal{L}_{token}$ is to ensure that the predicted fixation pack has the same length as the true fixation pack, with $v$ to represent the validity of the current fixation. Notably, due to the severe class imbalance between valid and invalid points, we adopt the $Focal Loss$~\cite{focalloss} to address this issue. 

\section{Experiments}
We train and evaluate ScanVLA on RefCOCO-Gaze dataset~\cite{lookhear}, which is currently the only substantial dataset available for ORSP task, encompassing 2,094 COCO~\cite{coco} images and 19,738 scanpaths.  
We evaluate ScanVLA from two aspects: scanpath similarity ($SS$, $SS_{pack}$, $FED$, $FED_{pack}$) and fixation saliency ($CC_{pack}$, $NSS_{pack}$)~\cite{lookhear}. In general, higher values of $SS$, $SS_{pack}$, $CC_{pack}$ and $NSS_{pack}$, combined with lower values of $FED$ and $FED_{pack}$, indicate better similarity between the model-generated scanpaths and the ground-truth scanpath. We compare ScanVLA against the following baselines: 
Human, Random Scanpath, OFA~\cite{ofa}, Chen \textit{et al.}~\cite{vqachen}, Gaze-ref\cite{gazeformer}, Gaze-cat\cite{gazeformer}, ART~\cite{lookhear}.

\subsection{Implementation Details} 
In our experiments, the maximum number of predictable fixations per pack is set to $L_p =4$, as the proportion of packs with four or less fixations in the RefCOCO-Gaze dataset already reaches 98.08\%. All experiments were conducted on four NVIDIA GeForce RTX 3090 GPUs, with each GPU processing a batch size of 1 and an accumulative gradient count of 8. We utilized the AdamW~\cite{adamw} optimizer and adopted a two-stage learning rate schedule, consisting of a LinearLR warmup phase followed by a CosineAnnealingLR decay phase.
We performed LoRA fine-tuning to LLM component, with LoRA rank set to 16 and alpha set to 32. To adhere to the same experimental settings as prior works~\cite{lookhear, gazeformer}, we deliberately employed images of the same size $(H = 312, W = 520)$ as input.

\subsection{Quantitative Comparison}
Table \ref{table: main} presents the quantitative evaluation results of our method ScanVLA and competitive baselines. As shown in Table \ref{table: main}, our model constantly and significantly outperforms all existing approaches across all metrics.
Notably, our method attains results that are closest to human performance and even surpasses human performance in several metrics. This is because in scanpath prediction, researchers collect multiple real human scanpaths for each image. Human performance shows the average similarity (or distance) between these real human scanpaths. Considering that human scanpaths are influenced not only by objective visual attention mechanisms but also by the subjective factors of different individuals, therefore, human performance is only a reference upper limit in scanpath prediction. Our method surpassing human performance indicates that our predicted scanpaths are more objective and contain fewer subjective components.

\begin{table}[t]
\centering
\setlength{\tabcolsep}{0.3mm}
\small
\caption{\textbf{Performance of ScanVLA and other baselines on RefCOCO-Gaze test set}.}
\vspace{-6pt} 
\begin{tabular}{lcccccc} 
\toprule
                     & SS↑            & SSpack↑        & FED↓           & FEDpack↓       & CCpack↑        & NSSpack↑        \\ 
\hline
Human                & 0.400          & 0.317          & 6.573          & 1.278          & 0.283          & 3.112           \\ 
\hline
Random               & 0.189          & 0.133          & 17.735         & 3.005          & 0.094          & 1.689           \\
OFA~\cite{ofa}                  & 0.216          & 0.170          & 17.084         & 2.901          & 0.174          & 2.175           \\
Chen \textit{et al}.~\cite{vqachen} & 0.299          & 0.188          & 8.309          & 1.507          & 0.159          & 1.557           \\
Gazeformer-ref~\cite{gazeformer}       & 0.269          & 0.194          & 6.788          & 1.286          & 0.208          & 3.006           \\
Gazeformer-cat~\cite{gazeformer}      & 0.269          & 0.189          & 6.841          & 1.327          & 0.204          & 2.932           \\
ART~\cite{lookhear}          & 0.359          & 0.292          & 6.371          & 1.143          & 0.280          & 3.478           \\ 
\hline
Ours (1B)                 & 0.389 & 0.328 & 5.986 & 1.099 & 0.318 & 3.810  \\
Ours (2B)                 & \textbf{0.407} & \textbf{0.344} & \textbf{5.864} & \textbf{1.077} & \textbf{0.354} & \textbf{4.203} \\
\bottomrule
\end{tabular}
\label{table: main}
\end{table}

\begin{figure}[t]
\centering
\includegraphics[width=0.9\columnwidth]{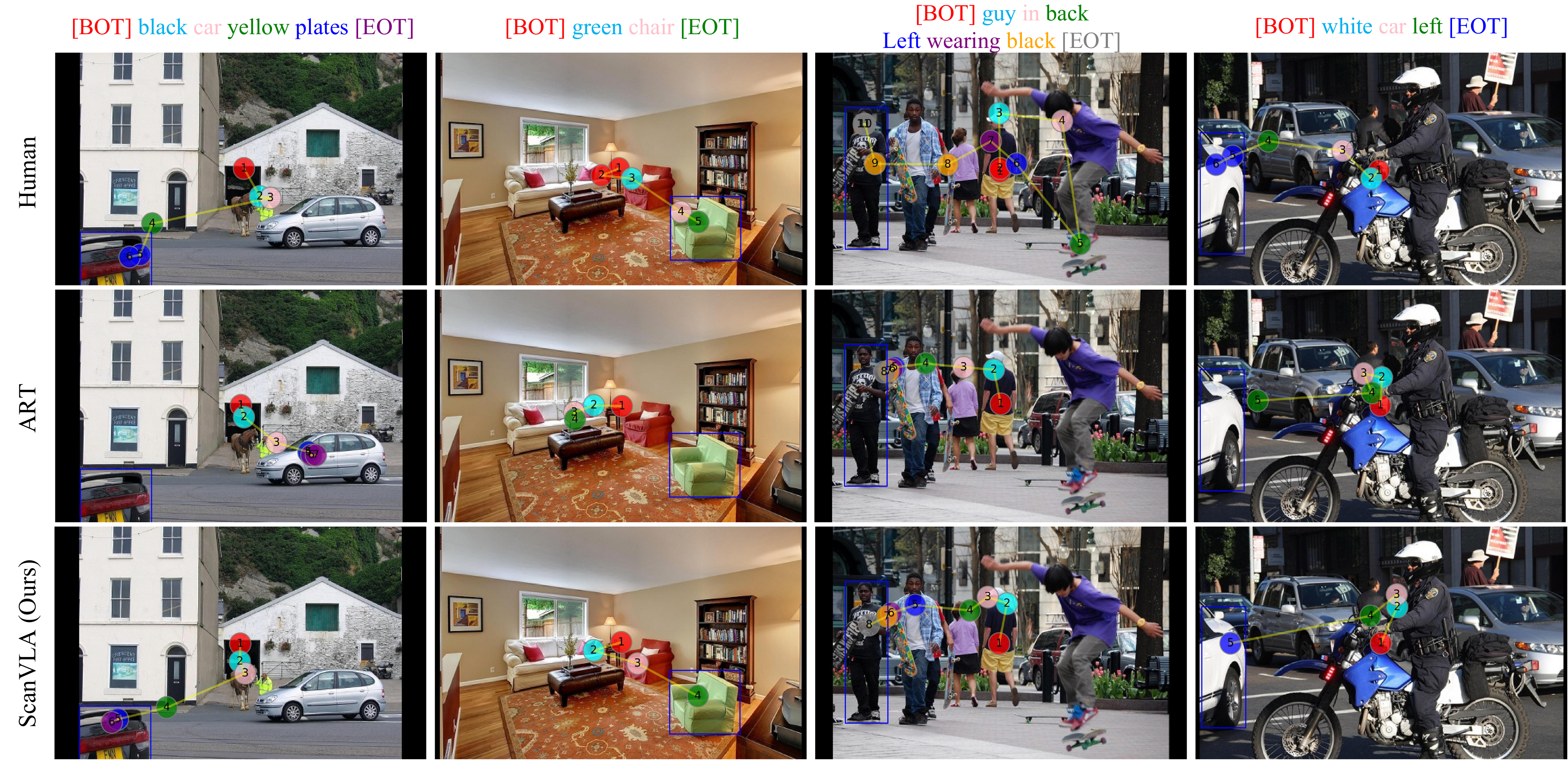} 
\caption{Qualitative comparison among our model, ART\cite{lookhear}, and the human groundtruth. 
}
\label{fig: Qualitative}
\vspace{-9pt}
\end{figure}

\subsection{Qualitative Comparison}
Figure \ref{fig: Qualitative} shows the qualitative comparison results among our model, the current SOTA model ART\cite{lookhear}, and the human groundtruth. 
As we can see, 1) ART confuses category-similar objects with the exact referent (first two columns, e.g., car, chair) due to visual-semantic mismatches; 2) even with correct target identification (last two columns), its fixations miss the bounding box, indicating positional insensitivity.
Compared with ART, our fixations target salient, human-relevant areas. These results confirm ScanVLA’s strengths in referential understanding, fixation accuracy, and saliency.

\subsection{Ablation Study}

To verify the effectiveness of our model architecture, we conducted comprehensive ablation studies on the 1B-sized ScanVLA from the following perspectives: 
First, we removed the frozen Segmentation LoRA component in LLM (denoted as `w/o Segment'). 
Second, we removed the Scanpath LoRA component, restricting training to only the HESD (denoted as `w/o Scanpath'). 
Third, We replace the proposed HESD with a simple linear (denoted as `w/o HESD').
Fourth, we omitted the text prediction loss $\mathcal{L}_{txt}$ to examine whether the auxiliary task of predicting the next word is useful (denoted as `w/o $\mathcal{L}_{txt}$'). 
Fifth, we replace the $L_1$ loss function with the $L_2$ loss function for $\mathcal{L}_{xy}$ (denoted as `$\mathcal{L}_{xy}$ with $L2$'). 
Sixth, instead of concatenating the historical fixation information output by the History GRU with the text hidden representations $\mathcal{H}_j, j\in [0,1,..,L+1]$ from the output of the VLM, we add it to the text representations $f_t$ at the input end of the LLM (denoted as `early fusion').

\begin{table}[t]
\centering
\setlength{\tabcolsep}{0.3mm}
\small
\caption{\textbf{The Result of Ablation.}}
\vspace{-6pt} 
\begin{tabular}{lcccccc} 
\toprule
                      & SS↑                  & SSpack↑              & FED↓                 & FEDpack↓             & CCpack↑              & NSSpack↑              \\ 
\hline
w/o Segment LoRA   & 0.390                & 0.326                & 6.026                & 1.107                & 0.317                & 3.795                 \\
w/o Scanpath LoRA     & 0.382                & 0.315                & 6.155                & 1.134                & 0.303                & 3.612                 \\
w/o HESD   & 0.382              & 0.325                & 6.016                & 1.104                & 0.314                & 3.793                 \\  
w/o $\mathcal{L}_{txt}$ & \textbf{0.394}                & 0.325                & 5.992                & 1.104                & 0.317                & 3.741                 \\
$\mathcal{L}_{xy}$ with $L2$   & 0.375              & 0.308                & 6.157                & 1.124                & 0.294                & 3.578                 \\  
early fusion   & 0.364              & 0.287                & 6.338                & 1.163                & 0.261                & 3.351                 \\  
\hline
Ours (1B)                  & 0.389       & \textbf{0.328}       & \textbf{5.986}       & \textbf{1.099}       & \textbf{0.318}       & \textbf{3.810}        \\
\bottomrule
\end{tabular}
\label{table: ablation}
\vspace{-19.65pt}
\end{table}

Table \ref{table: ablation} demonstrates that `Ours' achieves the best overall performance. Specifically, the performance degradation in `w/o Segment' and `w/o Scanpath' validates the importance of the two LoRA modules.
The results for 'w/o HESD' demonstrate the effectiveness of our proposed HESD module.
The results for `w/o $\mathcal{L}_{txt}$' demonstrate that the text prediction loss serves as a beneficial auxiliary objective.
Meanwhile, `$\mathcal{L}_{xy}$ with $L2$' indicates that $L1$ loss is more suitable, likely due to outliers in the dataset. Finally, the inferior results of `early fusion' suggest that integrating historical fixation information at the input level may impair visual-semantic alignment, leading to suboptimal performance. 

\section{Conclusion}
In this paper, we propose ScanVLA, a novel approach for the ORSP task. To the best of our knowledge, it is the first attempt to leverage VLMs for scanpath prediction, effectively addressing the feature alignment problem in multimodal information fusion.
Furthermore, to enhance ScanVLA’s capability of perceiving fine-grained positional information and referential targets, we not only design a HESD that directly takes the positional information of historical fixations as input, but also incorporate a frozen Segmentation LoRA as an auxiliary module to improve the model's ability to perceive and localize referential objects.
A trainable Scanpath LoRA is also used to capture ORSP-specific knowledge. Extensive experiments demonstrate that ScanVLA constantly and significantly outperforms existing SOTA models in various evaluation metrics,
exhibiting superior performance in understanding referring expressions, generating accurate fixations, and capturing salient regions.

\bibliographystyle{ACM-Reference-Format}
\bibliography{main}

\end{document}